# Embedding generation for text classification of Brazilian Portuguese user reviews: from bag-of-words to transformers

Frederico Dias Souza[1] · João Baptista de Oliveira e Souza Filho[1]

## Abstract
Text classification is a natural language processing (NLP) task relevant to many commercial applications, like e-commerce and customer service. Naturally, classifying such excerpts accurately often represents a challenge, due to intrinsic language aspects, like irony and nuance. To accomplish this task, one must provide a robust numerical representation for documents, a process known as embedding. Embedding represents a key NLP field nowadays, having faced a significant advance in the last decade, especially after the introduction of the word-to-vector concept and the popularization of Deep Learning models for solving NLP tasks, including Convolutional Neural Networks (CNNs), Recurrent Neural Networks (RNNs), and Transformer-based Language Models (TLMs). Despite the impressive achievements in this field, the literature coverage regarding generating embeddings for Brazilian Portuguese texts is scarce, especially when considering commercial user reviews. Therefore, this work aims to provide a comprehensive experimental study of embedding approaches targeting a binary sentiment classification of user reviews in Brazilian Portuguese. This study includes from classical (Bag-of-Words) to state-of-the-art (Transformer-based) NLP models. The methods are evaluated with five open-source databases with pre-defined data partitions made available in an open digital repository to encourage reproducibility. The Fine-tuned TLMs achieved the best results for all cases, being followed by the Feature-based TLM, LSTM, and CNN, with alternate ranks, depending on the database under analysis.

## 1 Introduction

*Text Classification* is a fundamental NLP field that aims to automatically label textual units like sentences, paragraphs, or documents. It may target sentiment analysis, topic analysis, question answering, and natural language inference tasks. Particularly, sentiment analysis, also called opinion mining, envisages to infer people's opinions expressed in textual data, predicting if they are positive or negative [1–3].

Opinions may significantly influence human behavior. Nowadays, every company or brand struggles to be aware of individual and collective opinions about themselves that may contribute to leveraging their business [4]. For instance, streaming services are often interested in discovering users' opinions about a movie to feed their recommendation systems, and e-commerce platforms desire to know customers' impressions about offered products to improve the user experience.

With the growing volume of data automatically generated by the interaction of users with various digital platforms, opinions can be automatically mined through analyzing their reviews, thus not requiring specific surveys or interviews. For this reason, sentiment analysis over user reviews is of particular importance in many companies' decision-making, leading to the current demand for large-scale algorithms to perform this task.

Making the computers extract strategic information from a text excerpt relevant to solving a NLP task might be

✉ Frederico Dias Souza
fredericods@poli.ufrj.br

João Baptista de Oliveira e Souza Filho
jbfilho@poli.ufrj.br

[1] Electrical Engineering Program, Federal University of Rio de Janeiro, Rua Moniz Aragão, Rio de Janeiro 21941-594, Brazil

challenging. The first step usually involves transforming the texts into usable computational representations, such as vectors, tensors, graphs, or trees [5]. The process of generating a vector representation to a word or document is known as embedding. It is one of the most critical aspects for the automatic inference of sentiments from texts, representing the central aspect of this work.

NLP solutions exploiting Machine Learning models have been existing for a long time. However, in the last decade, a brand of algorithms popularly known as Deep Learning achieved impressive results, a process started with the Convolutional Neural Networks (CNNs) and Recurrent Neural Networks (RNNs) [5]. More recently, the Transformers architecture revolutionized the NLP area, leading to the Transformer-based Language Models (TLMs), responsible for redefining the state-of-the-art of many tasks. TLMs have the distinguishing characteristic of processing large portions of texts in parallel in a much deeper semantic manner than previous NLP models, one of the primary reasons for their performance. Moreover, such models have the additional benefit of addressing a range of NLP problems through transfer learning.

Regarding sentiment analysis, Li et al. [6] proposed the emotional recurrent unit (ERU), a RNN variant dedicated to conversational sentiment analysis, as a simple alternative to complex Deep Learning architectures often proposed as solutions to this task. Envisaging to better explain the feelings predicted by an AI system, Cambria et al. [7] developed a neurosymbolic AI framework, combining classic approaches, such as knowledge graphs, with deep learning. Moreover, based on knowledge graphs and in track with the recent trend of exploring graph neural networks (GNNs) in NLP [8], Liang et al. [9] proposed a GNN architecture that incorporates affective information to improve textual dependency graphs.

The sentiment analysis field in Brazilian Portuguese disposes of several databases, pre-trained resources, and models, most of them released in recent years, as can be seen in Pereira's work [10] and *Opinando* project [11], the latter led by the University of São Paulo. However, to our best knowledge, the literature lacks studies comparing the embeddings generated by the current state-of-the-art models (Transformed-based) with more consolidated approaches, such as Bag-of-Words, as well as well-established Deep Learning models (CNNs and RNNs).

## 1.1 Objectives and contributions

Due to the growing importance of inferring users' opinions in large amounts of data and motivated by the lack of experimental studies including more recent NLP models, this work aims at conducting a comprehensive experimental study of embedding alternatives to the sentiment classification task for Brazilian Portuguese texts, contemplating from traditional and basic options up to state-of-the-art models, in five open-source databases to ensure generality and reproducibility of the findings.

Although some works [12, 13] emphasize the importance of including neutrality in the sentiment analysis, we decided here to focus only on the positive and negative reviews, disregarding those neutral, thus assuming a binary sentiment analysis task. The rationale is that such classes represent a more direct and unambiguous user feedback in e-commerce transactions, often requiring a more effective response from the customer satisfaction services, a central aspect in some future applications aimed by us, in opposition to the still relevant but fuzzy neutral category.

In the following, there is a brief summary of the main paper objectives and contributions:

1. Provide a comprehensive study of feature generation techniques for text classification in Brazilian Portuguese, from those based on corpus statistics (TF-IDF) to transfer-learning approaches, including word embedding strategies and Transformer-based Language Models. This study also covers intrinsic embeddings generated by an end-to-end optimization of models like CNNs, RNNs, and Fine-tuned TLMs.

2. Evaluate the predictive performance of Transformer-based Language Models available for Brazilian Portuguese. Despite the recent advances in the NLP area, open-source models in this language have emerged only recently. Therefore, their systematic evaluation is lacking, especially regarding the text classification task. Thus, this study includes three multilingual and seven Portuguese TLMs.

3. Contribute to practitioners when choosing a feature extraction model for text classification, providing some guidelines relative to practical compromises between the predictive performance and the computational resources demanded by some state-of-the-art models.

## 1.2 Organization

This work is organized as follows. First, Sect. 2 contextualizes the models evaluated. Then, Sect. 3 describes all the steps involved in the evaluation pipeline, detailing the experiments and how the models were trained and evaluated. Section 4 discusses the results, critically analyzing the influence of the algorithm's hyperparameters and architectures. Finally, Sect. 5 summarizes the main findings of this work.

## 2 Theoretical background

This section briefly contextualizes the Machine Learning models covered in this work, grouping them into three families: Bag-of-Words (BoW), Classical Deep Learning (CDL), and Transformer-based Language Models, thus in increasing order of complexity and innovation.

### 2.1 Bag-of-words

The most basic feature-based approach for text classification is the Bag-of-Words. It produces embeddings that ignore the order, context, and semantic relations between words in a corpus [1]. Usually, BoW encompasses statistics extracted from the corpus, such as the word frequency or the term frequency-inverse document frequency (TF-IDF), or even aggregated pre-trained word vectors, as discussed below.

#### 2.1.1 TF-IDF

The TF-IDF is a document embedding approach based on the frequency of occurrence of each word into a collection of $N$ documents (corpus) [14]. The dimensionality of a document vector produced by the TF-IDF is equal to the vocabulary size, wherein the $i$-component of the vector related to the $j$th document vector is given by

$$w_{ij} = tf_{ij}\ idf_i; \quad idf_i = log\left(\frac{N}{df_i}\right), \tag{1}$$

where $tf_{ij}$ is the frequency of occurrence of the $i$th word into the $j$th document, while $idf_i$ denotes the inverse of the ratio between the number of documents containing the $i$th word, denoted as $df_i$, and the total number of documents $N$.

This embedding often results in very sparse vectors, favoring frequent words of the same document and penalizing those present in many documents [15]. The vocabulary may also include sequences of words with an arbitrary length $n$, referred to as n-grams, in addition to single document words. To constrain the vocabulary size, one may just consider taking the most frequent n-grams or can exploit Chi-squared and ANOVA F value tests [16]. Singular Value Decomposition (SVD) may also be used for deriving low-dimensional representations over TF-IDF embeddings. This combined approach is known as Latent Semantic Analysis (LSA) [17].

#### 2.1.2 Bag-of-word vectors

Pre-trained word embeddings are the most classic transfer-learning strategy in NLP. Such vectorial representations are learned in a large corpus, leading to a semantic vector space wherein words with a similar meaning are mapped to geometrically close vectors [1]. Word2vec [18] was the pioneer semi-supervised neural framework for learning this embedding modality. FastText [19] is a remarkable advance to Word2vec, representing each word by a bag of character n-grams to ensure an attractive balance between the predictive performance and the vocabulary size.

The most basic approach when generating document embeddings over word embeddings consists of averaging all vectorial representations from the words that integrate a document, a procedure commonly known as Bag-of-Word Vectors Averaged (avgBoWV). This straightforward approach assigns the same importance to all words in a corpus, irrespective of their relative relevance. Singh et al. [20] reported promising results when using the values of $idf_i$ (Eq. 1) as word weight factors, referred to as IDF Weighted Bag-of-Word Vectors (idfBoWV).

### 2.2 Classical deep learning

The state-of-the-art in NLP [1] is currently represented by Deep Learning architectures, which are neural-based models able to extract complex patterns in large-scale databases, despite exhibiting a slow and challenging training phase. As compared to BoW models, one of the major differences between DL and classical approaches resides in automatically extracting the word embeddings necessary for solving a given NLP task [2].

#### 2.2.1 Long short-term memory networks

Typically, RNNs access the mutual inter-dependencies between words in a sentence more effectively than feed-forward neural networks or BoW models, thus better capturing the corpus context [1]. Long Short-Term Memory (LSTM) is a popular RNN variant proposed by Hochreiter and Schmidhuber [21]. It successfully mitigates the vanishing and the exploding gradient issues commonly faced with standard RNNs [1]. In the context of text classification, some applications of LSTM include a multi-task learning framework [22], a bidirectional LSTM followed by a 2D max-pooling operation [23], and short text sentiment classification using pre-trained word embeddings [24, 25].

#### 2.2.2 Convolutional neural networks

RNNs are popular in NLP applications requiring sequential processing due to their time-dependent processing skills. However, CNNs can outperform RNNs when dealing with local and position-invariant patterns, like phrases expressing a particular sentiment ("I like") or a topic ("endangered species"), thus may constitute a handy tool for text classification [1, 26]. Kim [27] reported experiments in sentence classification exploiting fine-tuned and non-fine-tuned pre-trained Word2vec vectors and a single-layer CNN. Zhang and Wallace [28] analyzed effects from CNN design settings in the performance of a sentence classification model.

## 2.3 Transformer-based language models

Transformers [29] is a revolutionary Deep Learning architecture based on the self-attention mechanism that turned viable the parallel training of highly complex NLP models over massive data [30].

BERT (Bidirectional Encoder Representations from Transformers) is a distinguishing Transformed-based model proposed by Devlin et al. [31] with two design options: pre-trained (or feature-based) and fine-tuned. In the first, the model outputs define feature vectors that may be combined to provide a document embedding, and only the Machine Learning model on the top is trained. In the second, the whole architecture is trained in an end-to-end fashion using annotated data and targeting a specific task. The combination of the tokens produced by the BERT model may also lead to expressive gains in Named-Entity Recognition, even without fine-tuning, as verified in our previous work [32].

BERT authors have also open-sourced a variant trained with more than 100 languages, including Portuguese, named m-BERT. Based on the robustly optimized BERT approach (RoBERTa) [33], it was developed a multilingual version trained with data from 100 languages named XLM-RoBERTa [34] (XLM-R). Regarding language-specific TLMs for Brazilian Portuguese, Souza et al. [35] have open-sourced the BERTimbau Base and Large, trained exclusively on brWaC [36] corpus, the most extensive open Portuguese corpus up to this date, reporting advances in the state-of-the-art of many NLP tasks. Similarly, Carmo et al. [37] have open-sourced the PTT5 model, a T5 (Text-to-Text Transfer Transformer) [38] model trained on the BERTimbau corpus, showing remarkable results.

This work evaluated only open-source TLMs for Portuguese released by the Hugging Face [39] initiative, an open-source NLP community. Brazilian Portuguese variants of the GPT (Generative Pre-trained Transformer) on this platform do not have papers attached but were included in the experiments of this work, such as the GPTNeo Small Portuguese[1] and GPorTuguese-2[2], corresponding to fine-tuned versions of the GPTNeo 125M[3] and GPT2 Small [40], respectively.

---
[1] www.huggingface.co/HeyLucasLeao/gpt-neo-small-portuguese.
[2] www.huggingface.co/pierreguillou/gpt2-small-portuguese.
[3] www.huggingface.co/EleutherAI/gpt-neo-125M.

## 3 The text classification pipeline

The general pipeline adopted for deploying the embeddings alternatives for text classification is depicted in Fig. 1. The first step includes collecting annotated user reviews and cleaning them, a fundamental step when dealing with free insertion texts. After, the document content is split into textual sub-units named tokens. Then, the tokens with lower semantic relevance may be removed, depending on the approach under analysis. After, a single feature vector is generated to feed the classification model.

Since this work focuses on obtaining meaningful features for text classification, the experiments considered the same classifier, represented by a simple Logistic Regression, to access the embeddings' performance, for all analyzed alternatives.

The following sections describe these steps in detail. Table 1 provides an overview of all experiments, and the column "Work contributions" points out those that are original.

### 3.1 Databases

Previous work [41] described the collection, pre-processing, mining, and organization of five open-source databases of user reviews in Brazilian Portuguese, which were released in a public repository[4]. Olist, B2W, and Buscapé datasets include product reviews from e-commerce platforms; UTLC-Apps, from *Google Play Store* applications, and UTLC-Movies has reviews of movies available at the Filmow platform. Table 2 summarizes some major database characteristics for the reader's convenience. Table 3 shows two samples of each dataset, one with positive and the other with negative polarity. One may note the much bigger size of some databases compared to others. Besides, the Buscapé and UTLC-Movies have the longest sentences. As a result, these databases may include excerpts with more semantic value, benefiting more complex models.

### 3.2 Text pre-processing and tokenization

Except in the case of TLMs, the documents were lowercased and had URLs and special characters removed. For the TF-IDF and LSA approaches, words were converted to the English alphabet, i.e., had the accents removed, and the occurrences of "ç" were replaced by "c." The models based on word vectors (Bag-of-Word Vectors, CNN, and LSTM) only underwent this process when a given word was not found in the corresponding embedding vocabulary.

---
[4] www.kaggle.com/datasets/fredericods/ptbr-sentiment-analysis-datasets.

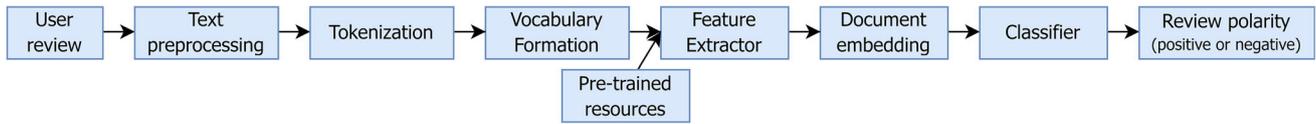

**Fig. 1** General pipeline considered for the text classification of user reviews in this work

**Table 1** Summary of the embedding approaches evaluated

| Model family | Model | Variations | Work contributions |
| --- | --- | --- | --- |
| BoW | TF-IDF | **Feature selection methods:** frequency, chi2, fvalue | Except for the "frequency" case, previously analyzed in [41], all the remaining are new. |
| | LSA | – | New |
| | avgBoWV | **Word vector dimension:** 50, 100, 300 | Analyzed in [41] |
| | idfBoWV | **Word vector dimension:** 50, 100, 300 | New |
| CDL | CNN | **Filter sizes:** [2], [2,3], [2,3,4], [2,3,4,5] **Feature map size**: 50, 100, 200, 400 | New |
| | LSTM | **Layers:** 1, 2 **Hidden size:** 64, 128, 256 **Pooling layer:** average, max, average and max | New |
| TLM | Feature-based TLM (FB TLM) | **Models:** see Table 4 **Aggregation types:** first, last, mean all, first + mean + std, mean + min+ max | Except for the BERT case analyzed in [32], the remaining are new. |
| | Fine-tuned TLM (FT TLM) | – | Analyzed in [32] |

**Table 2** Summary of the main characteristics of the databases analyzed in this work: the number of samples, the mean/median document length (in words), the vocabulary size (in words), the polarity distribution, and the percentage of corpus words covered in the NILC FastText word vector vocabulary [42]. Adapted from [32, 41]

| Database | Training/Validation/Test instances | Mean/Median document length (in words) | Vocabulary size (in words) | Polarity distribution (%) | FastText vectors coverage (%) |
| --- | --- | --- | --- | --- | --- |
| Olist | 30k / 4k / 4k | 7 / 6 | 3.272 | 70.0 | 96.5 |
| Buscapé | 59k / 7k / 7k | 25 / 17 | 13.470 | 90.8 | 92.9 |
| B2W | 93k / 12k / 12k | 14 / 10 | 12.758 | 69.2 | 94.4 |
| UTLC-Apps | 775k / 97k / 97k | 7 / 5 | 28.283 | 77.5 | 85.0 |
| UTLC-Movies | 952k / 119k / 119k | 21 / 10 | 69.711 | 88.4 | 91.1 |

**Table 3** Some examples of users' review (in Portuguese) for each database, with the corresponding polarity

| Database | User review (in Portuguese) | Polarity |
| --- | --- | --- |
| Olist | "O produto chegou no prazo combinado. Recomendo a loja" | 1 |
| | "Solicitei devolução! No site marcava um tamanho e veio menor" | 0 |
| Buscapé | "otimo pra quem quer uma foto com qualidade boa sem embaçados" | 1 |
| | "tv com imagem escura, e veio sem os itens da fabrica" | 0 |
| B2W | "Excelente produto, recomendo a família, e amigos. E a entrega foi rápida." | 1 |
| | "Ainda não recebi o produto, portanto não posso avaliá-lo!" | 0 |
| UTLC-Apps | "muito bom gostei de mais estão de parabéns" | 1 |
| | "horrível não consigo entrar no app e ninguém me responde no e-mail" | 0 |
| UTLC-Movies | "Obrigado Miyazaki por esse filme tão sutil, tão profundo e tão lindo..." | 1 |
| | "Um dos piores desfechos de filme que já vi na vida." | 0 |

This word vector match procedure is better explained in Sect. 3.4.

Regarding tokenization, Bag-of-Words and Classical Deep Learning models considered tokens with 2 up to 30 strings of letters separated by white spaces. Transformer-based Language Models considered raw input texts after being processed by the HuggingFace's AutoTokenizer[5]. The behavior of this AutoTokenizer class varies according to the model adopted, but it retains accents and capital letters, adding specific tokens whenever required by the corresponding transformer model. No further sentence processing (e.g., normalization, spell correction, named entity recognition) was conducted in this phase.

CDL models have the sentences padded to the 90% percentile of each database's number of tokens per document. Conversely, TLMs have the sentences padded to 60 tokens.

## 3.3 Vocabulary composition

TF-IDF experiments considered from 1 to 3 grams and different strategies to define the subset of n-grams from the corpus that must constitute the vocabulary. We did not consider n-grams beyond $n = 3$, since some preliminary tests pointed out a considerable increase in the computational time, especially for more extensive databases, such as UTLC-Movies, which was not followed by a significant gain on the classification accuracy.

All remaining methods considered only one gram. Words appearing less than five times in the corpus and the stop-words were removed from the BoW experiments. In the case of CDL models, all words to which pre-trained word vectors could not be found were ignored.

## 3.4 Pre-trained resources

The process of identifying the embedding of a given word was the following: the word, after being lower-cased and having possible special characters and URLs removed, is searched in the vocabulary list. If it is not found, it is converted to the English alphabet and searched again. If it remains not found, it is ignored. Table 2 depicts the percentage of words in each database that are covered by the FastText word vectors, signalizing a good embedding coverage in most cases (above 85%).

Transformer-based Language Models considered three multilingual and seven language-specific models, all publicly available for download on HuggingFace [39]. Table 4 summarizes the related embedding dimensionality and the number of parameters, aiming to provide some guiding information over the practical compromises between complexity and performance attained by these models.

## 3.5 Embeddings generation

Some practical issues related to the experiments reported in Table 1 are described in the following.

### 3.5.1 Bag-of-words

TF-IDF experiments with restricted-size vocabularies considered three alternatives of word selection: frequency, Chi-square test statistics ("chi2"), and ANOVA F value ("fvalue"), all available in the Scikit-learn [43] framework. BoW models exploited FastText word embeddings with 50, 100, and 300 dimensions, extracted from the NILC Word Embeddings Repository [42], and trained over a Brazilian Portuguese corpus. We considered both the arithmetic (avgBoWV) and the weighted (idfBoWV) average strategies for generating document vector embeddings.

### 3.5.2 Classical deep learning

A previous work [41] motivated us to assume 300-dimensional Fast Text word vectors and a logistic regression layer on the top of the CDL models. These experiments contemplated LSTM and CNN architectures, as detailed in the following.

1. LSTM: the architecture is exhibited in Fig. 2. It comprises one or two biLSTM layers, followed by a pooling layer, a dense layer with ReLU activation function (the output size is equal to the LSTM hidden size), and a linear dropout.
2. CNN: the architecture is based on [28] and depicted in Fig. 3. Roughly, it explores convolutional filters with different kernel sizes operating over the embeddings from the document's words. The resulting feature maps undergo a max-pooling operation, and the produced scalars are concatenated to feed a logistic regression layer with dropout.

### 3.5.3 Transformer-based language models

This work extends the results presented in [32] by considering a range of Transformer-based models beyond BERT. One of the major differences between BERT and alternative TLMs is its capability of accomplishing the Next Sentence Prediction (NSP) task. Besides, BERT has a token particularly dedicated to sentence classification (CLS). It is worth mentioning that this token can be included by fine-tuning, what would make these models more suitable for the text classification. However, this work restricts to analyze them in a feature-based fashion.

---

[5] www.huggingface.co/docs/transformers/model_doc/auto.

**Table 4** Summary of some Transformer-based Language Models' characteristics

| Model | Multi lingual | Embedding length | Parameters ($\times 10^6$) |
| --- | --- | --- | --- |
| PTT5 small | No | 512 | 60 |
| m-BERT | Yes | 768 | 110 |
| BERTimbau base | No | 768 | 110 |
| GPT2 small | No | 768 | 117 |
| XLM-Roberta base | Yes | 768 | 125 |
| GPTNeo small | No | 768 | 125 |
| PTT5 base | No | 768 | 220 |
| BERTimbau large | No | 1024 | 345 |
| XLM-Roberta large | Yes | 1024 | 355 |
| PTT5 large | No | 1024 | 770 |

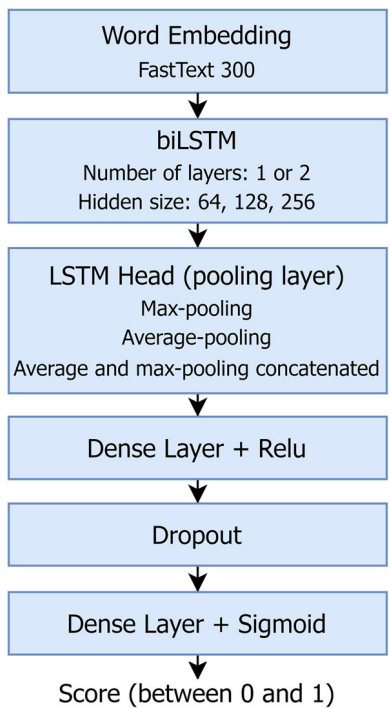

**Fig. 2** LSTM general architecture

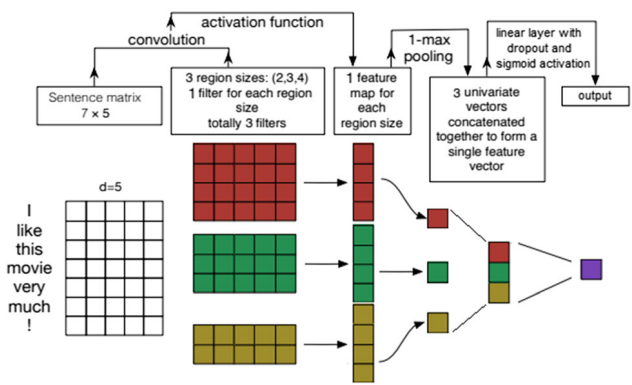

**Fig. 3** CNN general architecture. Adapted from [28]

As shown in Fig. 4, TLMs output a 3D-tensor whose number of rows is equal to the batch size, the number of columns is given by the number of tokens, and the depth corresponds to the embedding size. In this way, each column of this tensor represents a sequence of embeddings corresponding to some position-specific token of the documents integrating this batch. This factor has led us to evaluate the following mechanisms when generating the embedding for a given document:

1. **First**: the first token embedding;
2. **Last**: the last token embedding;
3. **Mean all**: the average of all token embeddings;
4. **First + mean + std**: concatenation of the embeddings of the first token, the average, and the standard deviation of the remaining tokens;
5. **Mean + min + max**: concatenation of the average, minimum, and maximum of all token embeddings.

Unlike BoW, which is based on static word embeddings, the document embeddings generated by TLMs are contextual, resulting in significant gains in the predictive performance of the models, as will be shown later. The TLM experiments that solely explore pre-trained models for document embedding generation will be referred here as Feature-based TLM (FB TLM). Conversely, those involving TLM fine-tuning, denoted here as Fine-tuned TLM (FT TLM), will be restricted to the BERTimbau Large model, motivated by findings from [32].

### 3.6 Classifiers

All document embeddings were evaluated using Logistic regression. In the case of BoW models and Feature-based TLMs, we adopted the *Scikit-learn* implementation with default parameters (the regularization factor seems not to significantly affect most of our experiments). Alternatively,

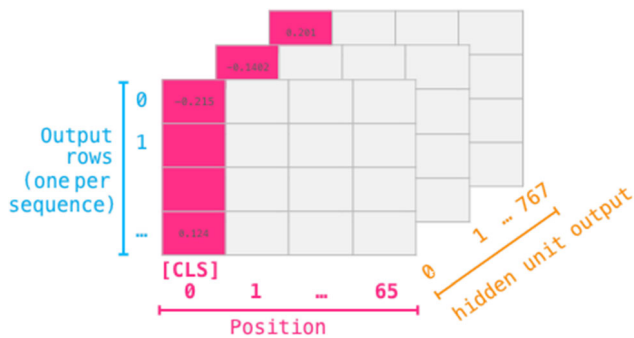

**Fig. 4** Example of BERT Base output. Adapted from [44]

Classical Deep Learning and Fine-tuned TLMs models have considered a neural implementation of this classifier.

### 3.7 Training procedure

For models not requiring hyperparameter tuning, the training phase was conducted with fused training and validation sets, and the performance metrics were derived with the test set. The hyperparameters of the remaining models were set using the cross-validation approach based on the validation set. After that, they were retrained with the training and validation sets concatenated.

### 3.8 Hyperparameter tuning

The process of hyperparameter tuning was restricted to Classical Deep Learning and Fine-tuned TLM, focusing on design aspects that typically have more impact on the model performance, like the learning rate and the number of training epochs. The hyperparameter search for Classical Deep Learning models considered several dropout (0, 5%, 10%, 15%, 20%, 25%, 30%, 35%, and 40%) and learning rate values ($5e-4$, $1e-3$, $5e-3$, $1e-2$). Inspired by [32], Fine-tuned TLM experiments explored two values for dropout (0 and 10%) and learning rates ($2.5e-5$ and $5e-5$).

### 3.9 Accessing models' performance

After identifying the models that show the best performance for each general embedding approach, we conducted a $k$-fold cross-validation experiment with $k = 10$, restricted to these cases due to computational reasons. Besides, we computed the average and standard deviation values of some performance metrics computed over the ten (pre-defined) test folds (as in [41]), as further described in Sect. 4.4. Finally, we submitted these results to statistical testing for a more rigorous analysis of the models' performance.

### 3.10 Computational resources

All models were implemented using the Google Colab Pro+ platform, a premium version of the Google Colab. Bag-of-Words models adopted a CPU with 16 GB of RAM. CDLs and TLMs exploited a P100 GPU with a vRAM of 16GB. For Feature-based TLMs, as we vectorized the entire corpus of each database before going through the classifier, we used Google Colab's high RAM environment option to store this large matrix, increasing the available RAM from 16 to 51GB.

## 4 Results and discussion

The figure of merit to select the best model for each embedding modality was the ROC-AUC. The rationale is that it is threshold invariant, summarizing the performance of a classifier for different operational settings, which may be established according to the risk associated with a wrong prediction. Due to our focus on e-commerce, such risks often vary according to the application domain, making the ROC-AUC an attractive performance metric.

For an overall model comparison, we also considered the Accuracy and F1-Score. The results grouped per embedding family are presented in the sequence, followed by a comparison with the best solutions identified in each case.

### 4.1 Bag-of-words

Figure 5 depicts results for the Bag-of-Words embeddings. In all cases, an increase in the dimensionality of the embeddings is followed by some gain in the model performance. The process of dictionary construction based on word frequency (the standard approach) has also been shown to be better than "chi2" and "fvalue" strategies, except in the case of the UTLC-Movies database. Up to 300 words, the LSA is the best strategy, being surpassed by the TF-IDF based on word frequency for embedding vectors with more than 1000 dimensions.

### 4.2 Classical deep learning

Tables 5 and 6 report results of the LSTM and CNN models, respectively. For the first, to most databases, except Buscapé, the differences in performance between the different architectures were very small (up to 0.3 percentage points). For the second, the models related to databases with short sentences, like Olist, B2W, and UTLC-Movies, seem not to be significantly affected by the size of the filters and the feature maps. On the other hand, considering databases with longer sentences (e.g., Buscapé and UTLC-Movies), the use of tuned hyperparameters allowed gains of at least one percentage point.

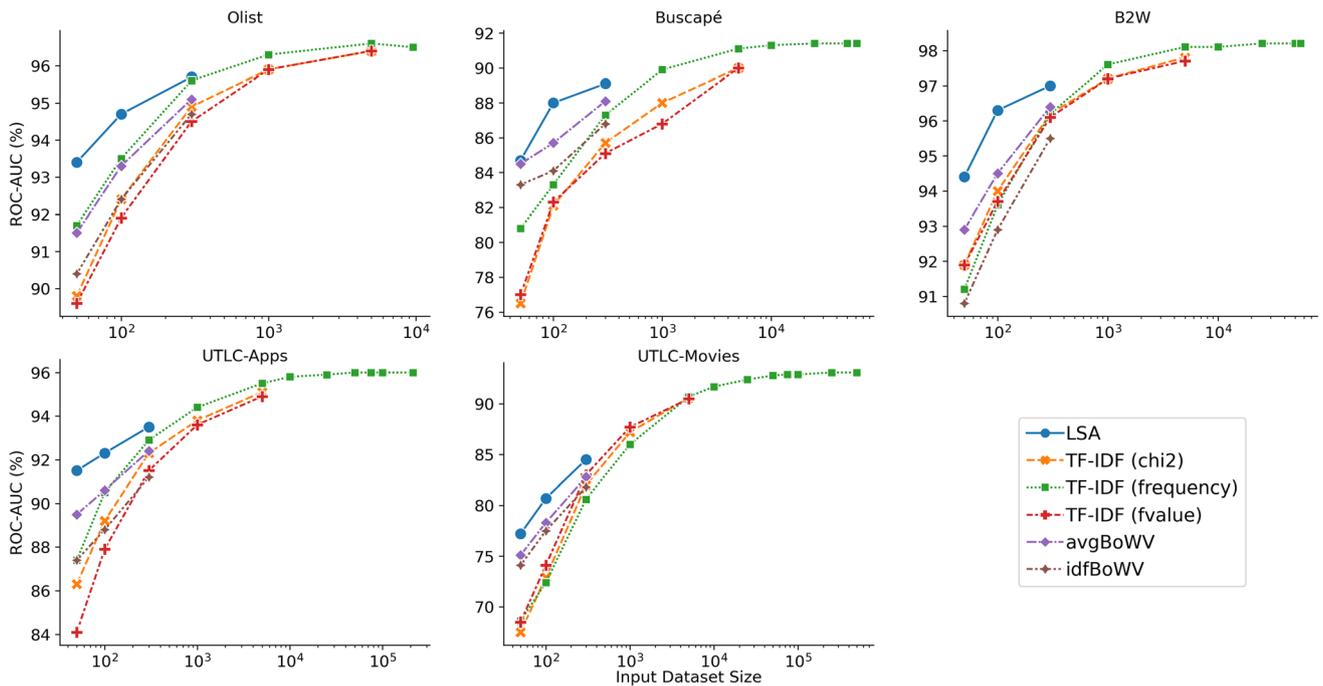

**Fig. 5** Values of ROC-AUC (%) for different dictionary sizes and BoW strategies (see text)

**Table 5** Values of ROC-AUC (%) obtained with LSTM models. The quantity "Delta" represents the difference between the best and the worst performance in each dataset. The highest value(s) of each database (column) is/are signalized in bold. The variables "N" and "HS" refer to the number of layers and the Hidden layer size, respectively

| N | HS | Pooling | Olist | Buscapé | B2W | UTLC apps | UTLC movies |
|---|---|---|---|---|---|---|---|
| 1 | 64 | Avg | **98.0** | 93.0 | **98.8** | 97.2 | 94.3 |
|   |    | Max | 97.9 | 92.9 | 98.7 | 97.2 | 94.4 |
|   |    | Avg‖Max | 97.8 | 92.8 | 98.7 | 97.2 | 94.4 |
| 1 | 128 | Avg | 97.7 | 92.2 | **98.8** | 97.2 | 94.3 |
|   |    | Max | 97.8 | 92.6 | **98.8** | 97.2 | **94.5** |
|   |    | Avg‖Max | **98.0** | 93.3 | **98.8** | **97.3** | **94.5** |
| 1 | 256 | Avg | 97.7 | 92.7 | **98.8** | 97.1 | **94.5** |
|   |    | Max | 97.9 | 93.0 | 98.6 | 97.2 | **94.5** |
|   |    | Avg‖Max | 97.7 | **93.5** | 98.7 | 97.1 | **94.5** |
| 2 | 64 | Avg | **98.0** | 93.0 | **98.8** | 97.2 | 94.3 |
|   |    | Max | 97.9 | 92.9 | 98.7 | 97.2 | 94.4 |
|   |    | Avg‖Max | 97.8 | 92.8 | 98.7 | 97.2 | 94.4 |
| 2 | 128 | Avg | 97.8 | 93.3 | 98.7 | 97.1 | 94.3 |
|   |    | Max | 97.9 | 93.4 | **98.8** | 97.2 | **94.5** |
|   |    | Avg‖Max | 97.8 | 93.1 | **98.8** | **97.3** | **94.5** |
| 2 | 256 | Avg | 97.7 | 92.7 | **98.8** | 97.1 | **94.5** |
|   |    | Max | 97.9 | 93.0 | 98.6 | 97.2 | **94.5** |
|   |    | Avg‖Max | 97.7 | **93.5** | 98.7 | 97.1 | **94.5** |
|   |    | Delta | 0.3 | 1.3 | 0.2 | 0.2 | 0.2 |

### 4.3 Feature-based transformer-based language models

This analysis considered ten Transformer models, five token aggregation modalities for document embedding, and five databases, as described in Sect. 2.3 and Table 2. After ranking the results for each database, which resulted in a list of integers from 1 to 50, an average rank for each TLM was computed to provide an overall performance index, regardless of the database and the aggregation modality. Table 7 reports the average rank obtained by each model, including the average time, the reserved vRAM, and the allocated vRAM required to process a batch with a size of 128. The allocated vRAM corresponds to the portion of GPU memory currently used, while the reserved vRAM is related to the cache memory allocated.

Table 6 ROC-AUC (%) for the CNN model. The quantity "Delta" represents the difference between the best and the worst performance in each dataset. The highest value(s) of each database (column) is/are signalized in bold

| Filter sizes | Feature map size | Olist | Buscapé | B2W | UTLC apps | UTLC movies |
| --- | --- | --- | --- | --- | --- | --- |
| 2 | 50 | 97.6 | 92.3 | 98.6 | 96.7 | 92.5 |
|   | 100 | 97.7 | 92.2 | 98.6 | **96.8** | 93.0 |
|   | 200 | 97.7 | 92.7 | 98.7 | **96.8** | 93.3 |
|   | 400 | **97.8** | 92.8 | **98.8** | **96.8** | 93.5 |
| 2, 3 | 50 | 97.6 | 92.1 | 98.7 | **96.8** | 92.7 |
|   | 100 | **97.8** | 92.6 | 98.6 | **96.8** | 93.3 |
|   | 200 | **97.8** | 92.7 | 98.7 | **96.8** | 93.5 |
|   | 400 | **97.8** | 92.9 | **98.8** | **96.8** | **93.8** |
| 2, 3, 4 | 50 | 97.7 | 92.6 | 98.6 | **96.8** | 93.1 |
|   | 100 | 97.7 | 92.8 | 98.7 | **96.8** | 93.4 |
|   | 200 | 97.7 | 93.0 | 98.7 | **96.8** | 93.5 |
|   | 400 | **97.8** | 93.0 | **98.8** | 96.7 | 93.6 |
| 2, 3, 4, 5 | 50 | 97.5 | 92.8 | 98.6 | **96.8** | 93.3 |
|   | 100 | 97.6 | 92.9 | 98.7 | **96.8** | 93.5 |
|   | 200 | 97.7 | 93.0 | 98.7 | **96.8** | 93.7 |
|   | 400 | 97.7 | **93.1** | 98.7 | 96.7 | **93.8** |
|   | Delta | 0.3 | 1.0 | 0.2 | 0.1 | 1.3 |

One may readily observe that increasing the number of parameters for models from the same family results in more predictive power. In other words, large model versions consistently outperform the corresponding base models. For the leading models, PTT5 Large is quite competitive with BERTimbau, followed by T5 and XLM-Roberta (XLM-R), representing significantly better alternatives than m-BERT, GPT2 Small, and GPTNeo Small.

We must stress that GPT models seem unsuitable for embedding generation in our task due to their language model structure based only on decoder blocks and unidirectional token processing, which is adequate for content generation.

Regarding the remaining parameters reported in Table 7, generally speaking, the reserved vRAM is correlated with the model size, while the allocated vRAM is with the embedding size. Therefore, these numbers shed some light on the practical compromises between predictive performance, computational time, and computational requirements observed to each model. As compared to the corresponding Base versions, the large models usually spend up to twice as much processing time and memory space. The XLM-Roberta obtained interesting results despite being multilingual, thanks to a cumbersome use of computational resources. PTT5 large is quite competitive with BERTimbau in terms of performance and processing time but it is significantly more memory hungry (almost twice). Therefore, BERTimbau Base is the model with the most attractive performance and computational trade-off.

Table 7 Average ranks, processing time, and quantities of reserved vRAM (GB) and allocated vRAM (GB) required to process a batch with size 128 for the Transformer-based Language Models. The process of rank calculation and the modalities of vRAM usage are discussed in the first paragraphs of Sect. 4.3

| Model | Average rank | Time (s) | Reserved vRAM | Allocated vRAM |
| --- | --- | --- | --- | --- |
| BERTimbau Large | 9.4 | 1.0 | 1.7 | 1.3 |
| PTT5 Large | 9.7 | 1.0 | 3.3 | 2.8 |
| BERTimbau Base | 16.8 | 0.5 | 0.8 | 0.4 |
| T5 Base | 17.2 | 0.5 | 1.3 | 0.8 |
| XLM-R Large | 22.7 | 1.5 | 10.3 | 2.1 |
| XLM-R Base | 26.6 | 0.9 | 8.8 | 1.0 |
| PTT5 Small | 36.3 | 0.5 | 0.5 | 0.2 |
| GPT2 Small | 37.0 | 0.6 | 3.6 | 0.5 |
| m-BERT | 38.0 | 0.5 | 1.1 | 0.7 |
| GPTNeo Small | 45.4 | 0.6 | 3.4 | 0.5 |

Figure 6 depicts the performance of the top-seven TLMs for a better visualization of the results, considering different token aggregation modalities. We must stress that the good results obtained with the BERTimbau in the case to which only the first token embedding is considered for the classification task reflects how relevant is disposing of a preferential token for classification (CLS). Other models, such as RoBERTa, may also include this resource if they are retrained, which is out of the scope of this work. In summary, all modalities exploiting more than one token outperformed single-token solutions. Besides, the modality "first + mean + std" achieved the best results for a wide variety of methods and databases. Therefore, this feature fusing strategy is interesting for boosting TLM models not disposing of specific tokens for classification. Nevertheless, the BERTimbau base is the most attractive model for all aggregation modalities evaluated here.

### 4.4 Overall comparison

To allow a more rigorous comparison of the models, we conducted a 10-fold cross-validation experiment involving the best performing models of each algorithm family (Table 8), using folds previously defined to each database [41]. Table 9 reports the mean and standard deviation values for Accuracy, F1-Score, and ROC-AUC derived from the ten test folds to each model and database. Besides, over the ROC-AUC values, we performed Friedman's Chi-square test [45] to assess if the differences observed in methods' performance are statistically significant, assuming a significance level of 5%. Then, we conducted a subsequent post hoc Tukey test to identify pairs of methods which performed statistically different (only to the cases wherein the first test signalized an overall difference). According to the first test, we have $\chi^2(4) > 33.5$ and $p < 0.001$ for all datasets, confirming that the differences observed in the methods' performance are meaningful. Based on the second, we can establish the following relations between methods' performance and databases, in order of increasing performance:

1. **Olist**: BoW < CNN ≡ LSTM < FB TLM < FT TLM
2. **Buscapé**: BoW < CNN = LSTM ≡ FB TLM < FT TLM
3. **B2W**: BoW < CNN ≡ LSTM < FB TLM < FT TLM
4. **UTLC-Apps**: BoW < CNN < LSTM ≡ FB TLM < FT TLM
5. **UTLC-Movies**: BoW ≡ FB TLM < CNN < LSTM < FT TLM

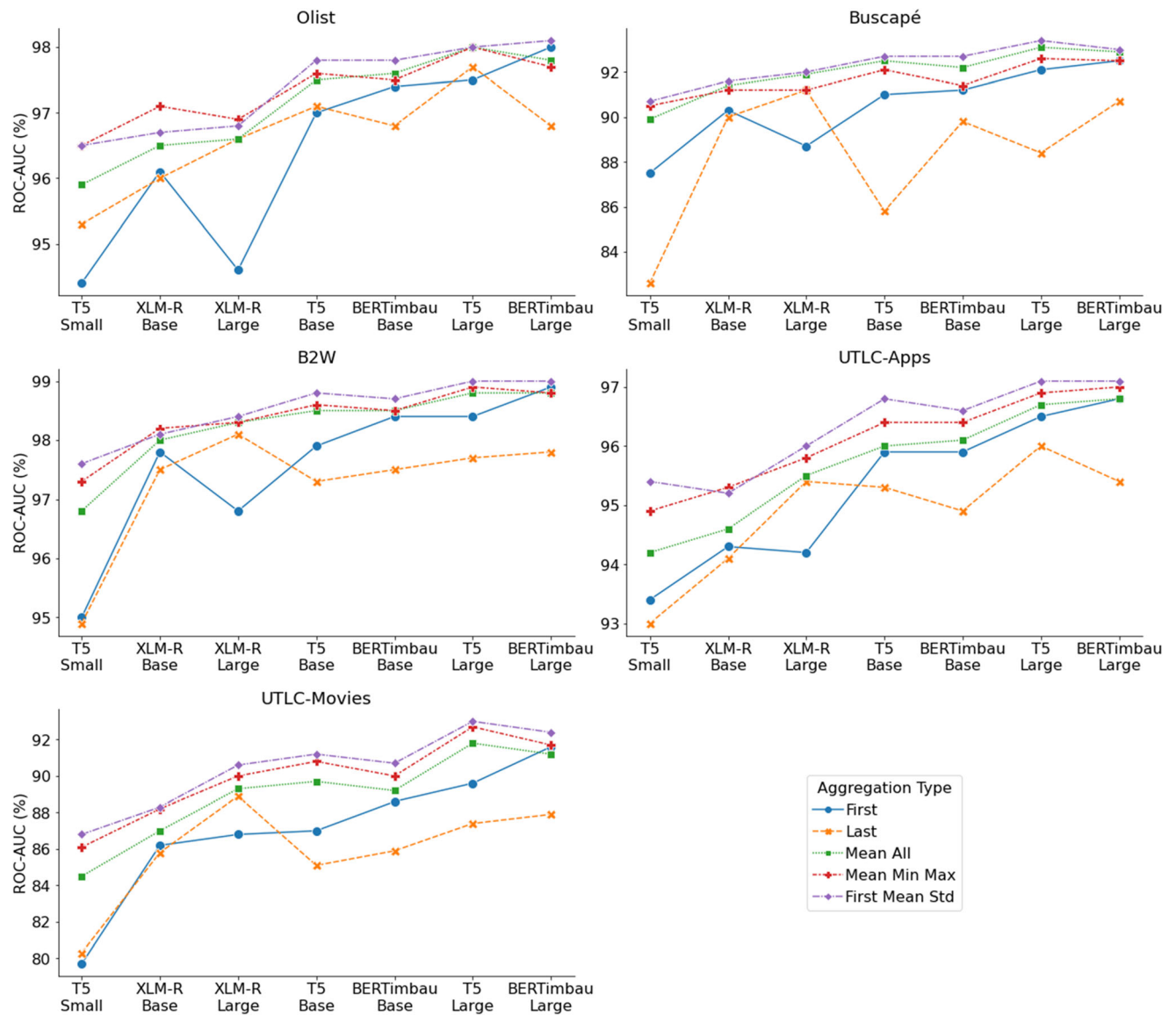

**Fig. 6** ROC-AUC (%) for each Transformer-based Language Model, embedding/aggregation modality, and database

Table 8 Hyperparameters of the best models identified for each database in terms of the vocabulary size (VS), learning rate (LR), dropout rate (DR), hidden size (HS), and Agg Type (aggregation type), as other factors

| Model | Param | Olist | Buscapé | B2W | UTLC apps | UTLC movies |
|---|---|---|---|---|---|---|
| TF-IDF | Feature selection | Frequency | | | | |
| | VS ($\times 10^3$) | 5 | 25 | 25 | 50 | 250 |
| CNN | Filters | 2,3 | 2,3,4,5 | 2,3 | 2,3 | 2,3 |
| | Feature size | 400 | | | | |
| | LR | $10^{-3}$ | | | | |
| | DR | 0.0 | 0.0 | 0.1 | 0.0 | 0.0 |
| | Epochs | 16 | 7 | 14 | 4 | 5 |
| LSTM | Layers | 1 | | | | |
| | HS | 128 | 256 | 128 | 128 | 128 |
| | Pooling | Average $\|\|$ Maximum | | | | |
| | LR ($\times 10^{-3}$) | 5 | 5 | 10 | 5 | 5 |
| | DR | 0.3 | 0.2 | 0.3 | 0.3 | 0.3 |
| | Epochs | 8 | 5 | 5 | 3 | 3 |
| FB TLM | Model | BERTimbau large | | | | |
| | Agg Type | First + mean + std | | | | |
| FT TLM | Model | BERTimbau large | | | | |
| | LR | $2 \times 10^{-5}$ | | | | |
| | DR | 0.1 | | | | |
| | Epochs | 1 | | | | |

Table 9 Mean and standard deviation of ROC-AUC (%), Accuracy (%), and F1-Score (%) values obtained with each model and database. The highest average values are signalized in bold

| Metric | Model family | Olist | Buscapé | B2W | UTLC-Apps | UTLC-Movies |
|---|---|---|---|---|---|---|
| Accuracy | BoW | 91.8 ± 0.2 | 94.8 ± 0.2 | 94.0 ± 0.3 | 92.3 ± 0.1 | 93.1 ± 0.0 |
| | CNN | 93.3 ± 0.6 | 95.7 ± 0.2 | 94.7 ± 0.6 | 93.1 ± 0.6 | 93.7 ± 0.1 |
| | LSTM | 93.4 ± 0.6 | 95.5 ± 0.2 | 94.4 ± 1.0 | 93.6 ± 0.1 | 94.0 ± 0.1 |
| | FB TLM | 94.7 ± 0.4 | 95.6 ± 0.2 | 96.1 ± 0.2 | 93.6 ± 0.1 | 93.2 ± 0.1 |
| | FT TLM | **95.3 ± 0.3** | **96.0 ± 0.1** | **97.0 ± 0.1** | **94.9 ± 0.1** | **95.2 ± 0.1** |
| F1-Score | BoW | 94.2 ± 0.2 | 97.2 ± 0.1 | 95.7 ± 0.2 | 95.1 ± 0.0 | 96.2 ± 0.0 |
| | CNN | 95.2 ± 0.4 | 97.7 ± 0.1 | 96.2 ± 0.4 | 95.6 ± 0.4 | 96.5 ± 0.0 |
| | LSTM | 95.3 ± 0.4 | 97.6 ± 0.1 | 96.0 ± 0.6 | 95.9 ± 0.1 | 96.6 ± 0.1 |
| | FB TLM | 96.2 ± 0.3 | 97.6 ± 0.1 | 97.2 ± 0.1 | 95.9 ± 0.0 | 96.2 ± 0.0 |
| | FT TLM | **96.6 ± 0.2** | **97.8 ± 0.1** | **97.8 ± 0.1** | **96.7 ± 0.1** | **97.3 ± 0.0** |
| ROC-AUC | BoW | 96.6 ± 0.3 | 91.9 ± 0.6 | 98.1 ± 0.1 | 96.1 ± 0.1 | 92.8 ± 0.1 |
| | CNN | 97.7 ± 0.1 | 93.3 ± 0.6 | 98.8 ± 0.1 | 97.0 ± 0.1 | 93.7 ± 0.1 |
| | LSTM | 97.6 ± 0.2 | 93.2 ± 0.6 | 98.8 ± 0.1 | 97.3 ± 0.1 | 94.5 ± 0.1 |
| | FB TLM | 98.0 ± 0.1 | 93.0 ± 0.6 | 99.0 ± 0.1 | 97.2 ± 0.1 | 92.9 ± 0.1 |
| | FT TLM | **98.4 ± 0.2** | **94.3 ± 0.4** | **99.4 ± 0.1** | **98.1 ± 0.1** | **96.1 ± 0.1** |

As expected, the Bag-of-Words is the worst, while the Fine-tuned BERT is the best in all cases. CNN and LSTM performed similarly for Olist, Buscapé, and B2W databases, while LSTM surpassed CNN in both UTLC databases. FB TLM is generally equivalent or superior to the Classical Deep Learning models, except to UTLC-Movies, a database with more samples, longer, and more complex sentences, to which models involving some training performed better. Considering practical applications, the BoW models, despite a lower predictive performance, are still attractive due to their implementation simplicity and low computational burden, not requiring the use of high-end GPUs. If one focuses on predictive performance, FB TLM represents an excellent intermediate alternative, as it generally performs better than the Classical Deep Learning models and has a simple implementation when open-source pre-trained models are used.

## 5 Conclusion

This work represents a comprehensive experimental study of document embedding strategies targeting text classification, including from classical to recently proposed Transformed-based models, the latter exploiting a transfer-learning paradigm. The main paper findings can be summarized as follows.

The simple TF-IDF approach outperformed more complex word vector aggregation strategies (avgBoWV and idfBoWV) with the increase in the embedding dimensionality, particular for more than 1000 dimensions, representing an attractive compromise between complexity and performance, especially for more straightforward classification tasks over low-complex semantic texts. Classical Deep Learning models, such as CNN and LSTM, seem to be more significantly affected by model hyperparameters only when dealing with databases having more complex and longer sentences, such as Buscapé and UTLC-Movies. As pointed out by the statistical tests, their performance tends to be superior to BoW models and inferior to Transformer-based Language Models. Regarding Transformer-based Language Models, increasing the model complexity usually leads to higher performance for the same architecture (Large vs Base). Models exclusively trained with Portuguese corpora like BERTimbau and PTT5 obtained the best results. Surprisingly, the multilingual XLM-Roberta has shown to be competitive with BERTimbau and PTT5.

Moreover, all Transformer-based models have benefited from the aggregation of tokens when generating document embeddings. From a practical point of view, the BoW model is convenient due to its implementation simplicity and reduced computational cost. In turn, the Feature-based TLM represents a solid alternative with an intermediate performance to application scenarios in which a higher predictive performance is required.

**Data availability** The databases used to train and evaluate the models are available at www.kaggle.com/datasets/fredericods/ptbr-sentiment-analysis-datasets.

## Declarations

**Conflict of interest** The authors declare that they have no conflict of interest.